# Assessing The Performance Bounds Of Local Feature Detectors: Taking Inspiration From Electronics Design Practices

Shoaib Ehsan[1], Adrian F. Clark[1], Bruno Ferrarini[1], Naveed Ur Rehman[2] and Klaus D. McDonald-Maier[1]

[1] 1School of Computer Science and Electronic Engineering, University of Essex, Wivenhoe Park, Colchester, UK
[2] Department of Electrical Engineering, COMSATS Institute of Information Technology, Islamabad, Pakistan
sehsan@essex.ac.uk

*Abstract* - **Since local feature detection has been one of the most active research areas in computer vision, a large number of detectors have been proposed. This has rendered the task of characterizing the performance of various feature detection methods an important issue in vision research. Inspired by the good practices of electronic system design, a generic framework based on the improved repeatability measure is presented in this paper that allows assessment of the upper and lower bounds of detector performance in an effort to design more reliable and effective vision systems. This framework is then employed to establish operating and guarantee regions for several state-of-the art detectors for JPEG compression and uniform light changes. The results are obtained using a newly acquired, large image database (15092 images) with 539 different scenes. These results provide new insights into the behavior of detectors and are also useful from the vision systems design perspective.**

I. INTRODUCTION

Consider designing an electronic system such as an integrated circuit. A designer cannot size the components regardless of the operating conditions and the productive process tolerance, which causes the components' parameters vary around their nominal value. The designers, indeed, need to know accurately the upper and lower performance bounds of the utilized electronic components in order to predict more easily the output of the system as a whole under different scenarios, such as large variations in temperature. The main motivating factor behind this approach is to make the designed system as much reliable as possible.

Now come back to the computer vision world and design a simple toy car tracking system with local feature detection as its primary stage while expecting only 20% uniform decrease in illumination. Looking at the repeatability results presented in [1] (which are widely considered the most comprehensive) for the Leuven dataset (which involves uniform changes in light) [2], MSER detector [3] appears to be the best option for achieving a reasonable value of repeatability (more than 60%) for this small transformation amount. Now consider two sample images (shown in Figure 1). which the designed vision system would encounter when deployed in the actual environment. The first image is the reference image and the second image has undergone 20% uniform decrease in light relative to the reference. Theoretically speaking, the feature detection unit (based on MSER) of the designed vision system would achieve high repeatability score for this negligible image transformation. As it turns out, MSER only manages a repeatability value of only 28.17% for the image pair shown, which is much less than what is expected of the feature detection unit and highlights its unreliable behavior – a stark contrast to the electronic system design example. If every component of an electronic system has known operating characteristics (or performance limits), the same should be for a vision system. In other words, for achieving the goal of reliability and effectiveness, the operating characteristics of every component in the vision system should be well known – something which is in line with the electronic system design practices but has not been done yet. Inspired by the good practices of electronic systems design, this paper addresses the problem of characterizing the upper and lower operative limits of local feature detectors, which form the initial stage of most vision systems today. In this work, we propose a generic framework for finding the operating and guarantee regions of local feature detectors under JPEG compression and light changes utilizing a large database of 15092 images involving 539 real world scenes. We base our framework on the improved repeatability rate introduced in [4].

The paper is organized as follows. Section II provides an overview of the related work in the domain of local feature detection evaluation. Section III proposes the generic framework for finding the operating and guarantee regions of a given local feature detector under a specific transformation. The newly acquired, large database of images is introduced in Section IV. The results are presented and discussed in Section V. Finally, the conclusions are given in Section VI.

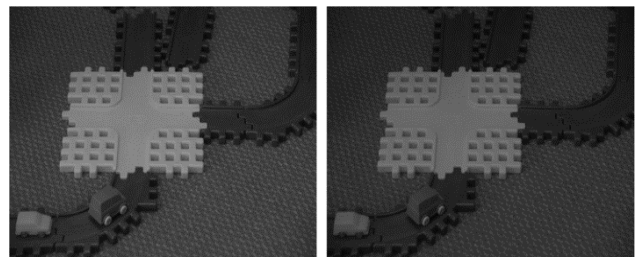

Figure 1. Two sample images; the left image is the reference image whereas the right image undergoes 20% uniform decrease in illumination.

## II. RELATED WORK

The literature on evaluation of local feature detectors is vast and has grown rapidly after the emergence of SIFT [5]. It is not possible to describe every such contribution here but an attempt has been made to mention all those developments which are considered important in this domain. Repeatability and information content are utilized as performance metrics in [6]. These two measures are also used in [7] for evaluating feature detectors in the context of image retrieval. The definition of repeatability was refined by [8] and used for evaluating six state-of-the-art local feature detectors in [1]. Improved repeatability measures are presented in [4]. The performance of local feature detectors is compared based on image coverage in [9] and [10]. Completeness of detected features is used as a performance metric in [11] for comparing state-of-the-art local feature detectors. In [12], the performance of detectors is evaluated under viewpoint, scale and light changes by using a large database of images with recall rate as performance measure.

## III. PROPOSED FRAMEWORK

Before discussing the details, it is worth stating that the proposed framework is based on the following principle: the ability to determine the upper and lower performance bounds of a given detector under some specific type and amount of image transformation — an idea borrowed from electronic systems design practice.

For achieving this objective, the framework utilizes the improved repeatability measure presented in [4] which provides results that are reliable and consistent with the actual performance of a wide variety of detectors across a number of well-established datasets

$$Repeatability = N_{rep}/N_{ref} \qquad (1)$$

where $N_{rep}$ is the total number of repeated points and $N_{ref}$ is the total number of interest points in the common part of the reference image.

Assuming the availability of a large image database involving a specific type of image transformation with known ground truth mapping between images and consisting of $n$ individual datasets with each having a different scene, the first component of the framework carries out the following steps:

Step 1: The repeatability scores are computed using Equation (1) for all images in every individual dataset (of the large image database) by taking the first image in each dataset which contains no transformation, as the reference. Assuming that the amount of image transformation is varied in $m$ discrete steps for every single dataset, $n$ values of repeatability are obtained for each discrete step. Let $A$ be the set of $m$ discrete steps representing specific transformation amounts

$$A = \{1, 2, 3, \ldots\ldots, m\} \qquad (2)$$

Let $B_k$ be the set of $n$ repeatability values at any one specific step $k$, where $k$ is an element of set $A$

$$B_k = \{b_{1k}, b_{2k}, \ldots\ldots, b_{nk}\} \qquad (3)$$

For example, if the image database consists of 539 different datasets (the number which will be used in the next few sections), each consisting of a sequence of 14 images, the values of n and m will be 539 and 14 respectively. In other words, there will be 539 values of repeatability available for each step of image transformation amount.

Step 2: For every discrete step $k$, the maximum value of repeatability is

$$P = \{\max(B_1), \max(B_2), \ldots\ldots, \max(B_m)\} \qquad (4)$$

The values of set $P$ are plotted against the corresponding image transformation amounts from set $A$ to obtain a curve which represents the upper bound of performance for the given detector with variation in the amount of transformation. This curve is named the max curve.

Step 3: For every discrete step $k$, the minimum value of repeatability is found to give

$$Q = \{\min(B_1), \min(B_2), \ldots\ldots, \min(B_m)\} \qquad (5)$$

The values of set $Q$ are plotted against the corresponding image transformation amounts from set $A$ to obtain a curve which represents the lower bound of performance for the given detector with the same variations in image transformation. This curve is named the min curve.

Step 4: For every discrete step $k$, the median value of repeatability is found

$$S = \{\text{median}(B_1), \ldots., \text{median}(B_m)\} \qquad (6)$$

The values of set $S$ are plotted against the corresponding image transformation amounts from set $A$ to obtain a curve which represents the typical performance for the given detector with variation in image transformation amount. This curve is named the median curve.

Step 5: By plotting the three curves together, the area between the max curve and the min curve is defined as the operating region of the detector. The detector is expected to produce repeatability scores that lie inside this region. A narrow operating region implies that the detector is stable and there is little variation between the maximum and minimum repeatability values that it can achieve for some specific amount of transformation. On the other hand, a large operating region indicates an unstable detector which may achieve high repeatability scores for some particular images but may fare poorly for others.

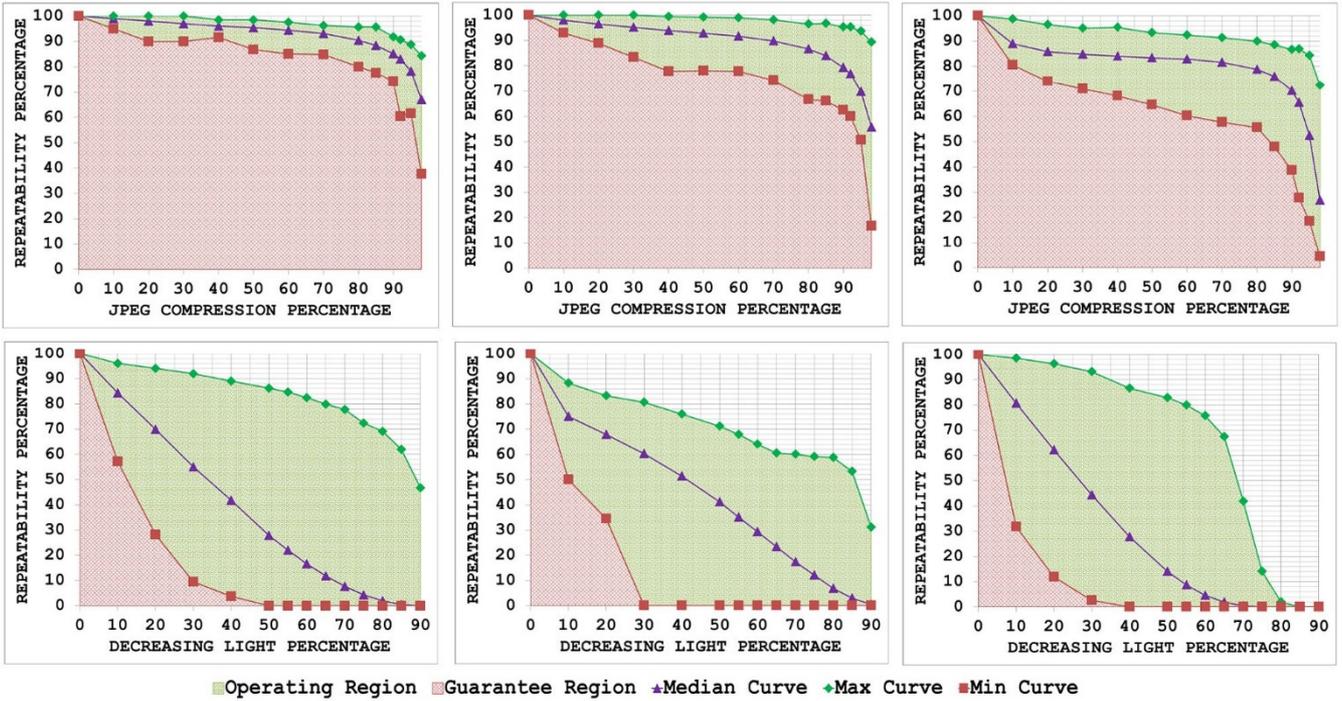

Figure 3. JPEG compression results utilising the proposed framework and the image database for Hessian-Laplace (top left), SURF (top centre) and SIFT (top right); Uniform light changes results for MSER (bottom left), IBR (bottom centre) and Harris-Laplace (bottom)

Step 6: The area under the min curve is defined as the guarantee region of the detector. Repeatability values achieved by the detector should never be as low so as to lie inside this region. A wide guarantee region shows that the detector manages to achieve reasonably high repeatability values for every input image with increasing amount of image transformation. Contrary to that, a small guarantee region implies that the detector performs poorly with increasing amount of image transformation.

## IV. THE IMAGE DATABASE

This section presents a newly acquired image database for finding the performance bounds of different local feature detectors. With 539 different scenes, the database contains 15092 images involving two image transformations, namely JPEG compression and uniform light changes. Some images from the image database are shown in Figure 2. To facilitate future research in this area, the image database is made available at [13]. In [1], the authors examined the performance of different local feature detectors on the basis of a single dataset, UBC [2] for JPEG compression ratios varying from 60% to 98%. Among the Oxford datasets [2], only Leuven, consisting of a sequence of six images, involves uniform changes in light. In [12], a large image database is presented to investigate the effect of light direction on the performance of feature detectors. However, the total number of scenes that have been used in that database is only 60. To investigate the behavior of local feature detectors by employing the framework proposed in the previous section reliably, a much larger database of images with variation in JPEG compression ratio and uniform light changes is required. Since there is no such resource available, this section presents a newly acquired database of images. The database consists of a dataset for each of the 539 scenes and transformation type: each dataset includes a reference image of the scene and several target images at increasing amounts of JPEG compression and uniform light changes. Each image in the database consists of 717 x 1080 pixels. The JPEG compression ratio is varied for every scene in 14 discrete steps from 0% to 98% (14 x 539 = 7546 images) and, similarly, the light brightness varies from 0% to 90% in the same number of steps (14 x 539 = 7546 images). The ground truth homography that relates any two images of the same scene with different imaging conditions in the presented database is a 3 x 3 identity matrix as both JPEG compression and uniform changes in light do not result in any geometric transformation.

## V. RESULTS

This section presents results for six state-of-the-art feature detectors for JPEG compression and uniform light changes utilizing the proposed framework and the large image database. These detectors include SIFT [5] and, SURF [14], Harris-Laplace, Hessian-Laplace [8], Intensity-based Regions (IBR) [15] and Maximally Stable Extremal Regions (MSER) [3]. These were chosen because they are scale- and rotation-invariant detectors and representative of a number of different approaches to feature detection [16]; also their implementations are widely available. Although the control parameters of these feature detectors can be varied to yield a similar number of interest points for all detectors, this approach has a negative effect on their repeatability and performance [8]. Therefore, authors' original programs (binary or source) have been utilized with parameters set to values recommended by them. The parameter settings used make these results a direct complement to existing evaluations. Due to space constraints, we are showing results for three detectors only in Figure 3 for each image transformation (Hessian-Laplace, SURF and SIFT for JPEG compression; MSER, IBR and Harris-Laplace for uniform light changes). The results in Figure 3 determine the upper and lower bounds of performance of detectors with varying JPEG compression ratio and uniform light changes, and then establish their operating and guarantee regions. Before discussing the results, it is worth stating that this appears to be the first attempt to do such a detailed analysis for these specific image transformations; there is no other work in the literature with which JPEG compression results can be compared to determine consistencies and contradictions. In [1], the authors have concluded that the six detectors under study are highly robust to uniform variations in illumination. As mentioned earlier, this deduction is based on a single dataset, Leuven [2]. The results presented here largely contradict those findings, showing that there is a rapid decline in the performance in the presence of uniform light changes. A similar performance degradation effect is observed in [12] while studying the behavior of feature detectors under changes in light direction. Therefore, the results presented here provide useful insight into the behavior of detectors under variations of JPEG compression and uniform light changes.

As evident from Figure 3, SURF performs well for increasing JPEG compression ratios up to 95% due to its wide guarantee region. It shows relatively poor stability only for the case when JPEG compression ratio is 98%. The narrow operating regions for Harris-Laplace (not shown in Figure 3 due to space constraints) and Hessian-Laplace demonstrate their stability to increasing JPEG compression. These two detectors also have wide guarantee regions, which indicate that they manage to achieve high repeatability scores even for large JPEG compression ratios. Although the performance of SIFT detector is reasonable, its operating region is wider than that of Hessian-Laplace and grows with increasing JPEG compression ratio. Moreover, the performance of SIFT may go to nearly zero depending upon the image content for 98% compression ratio (see the min curve of SIFT in Figure 3). MSER (not shown here) does perform well for some particular images with increasing JPEG compression ratio, the large operating region shows that its behavior is unstable. Even for small amounts of transformation, MSER fails to achieve high repeatability values for some images. IBR is more stable than MSER with increasing JPEG compression ratios as its operating region is smaller.

For uniform light changes, MSER and IBR have very large operating regions which indicate their unstable behavior in the presence of decreasing light. It is interesting to note that the min curve of MSER, the detector which is identified as the best for this specific image transformation in [1], reaches zero for only 50% uniform decrease in light. There is also rapid decline in the performance of SURF, Harris-Laplace and Hessian-Laplace detectors with decreasing light. The operating regions of these detectors are large and their guarantee regions are narrow, meaning that they may achieve high repeatability scores for some images but may fare poorly for others.

## VI. CONCLUSIONS

For designing reliable and more effective vision systems, this paper has presented a generic framework based on the improved repeatability measure [4]. The results based on this framework provide novel insights into the strengths and weaknesses of the detectors from a vision system design perspective. The results largely contradict the previous findings and provide new performance scores for the popular feature detectors under considered image transformations. These performance curves are more consistent with what experienced vision researchers expect and encounter.